# Lessons learned from deploying imaging AI with the open PACS-AI platform

**Authors:** Samuel Kadoury[1,2], Julie G. Hussin[3,4,5], Pascal Thériault-Lauzier[6], Laurent Létourneau-Guillon[1], Rob Lewis[7,8], Adam McArthur[9], Gordon J. Harris[7,10,11], Houda Bahig[12], Pierre-Luc Déziel[13,14], Jay Kshirsagar[6], Jacob L. Jaremko[9], Julien Cohen-Adad[4,15,16], Jacques Delfrate[17], Robert Avram[3,17]

**Except for the first author and the last (corresponding) author, authors are listed in random order (seed 20260812).**

1 Department of Radiology, Centre Hospitalier de l'Université de Montréal (CHUM) and Centre de recherche du CHUM (CRCHUM), Montréal, QC, Canada

2 Department of Computer and Software Engineering, Polytechnique Montréal, Montréal, QC, Canada

3 Department of Medicine, Montreal Heart Institute, Université de Montréal, Montréal, QC, Canada

4 Mila - Quebec AI Institute, Montréal, QC, Canada

5 Department of Medicine and Department of Biochemistry and Molecular Medicine, Faculty of Medicine, Université de Montréal, Montréal, QC, Canada

6 University of Ottawa Heart Institute, Ottawa, ON, Canada

7 Open Health Imaging Foundation, Boston, MA, USA

8 Radical Imaging, Bolton, MA, USA

9 Department of Radiology and Diagnostic Imaging, University of Alberta, Edmonton, AB, Canada

10 Department of Radiology, Massachusetts General Hospital and Harvard Medical School, Boston, MA, USA

11 Yunu, Inc., Cary, NC, USA

12 Radiation Oncology Department, Centre Hospitalier de l'Université de Montréal, Montréal, QC, Canada

13 Faculty of Law, Université Laval, Québec City, QC, Canada

14 Centre de recherche du CHU de Québec-Université Laval, Québec City, QC, Canada

15 NeuroPoly Lab, Polytechnique Montréal, Montréal, QC, Canada

16 Functional Neuroimaging Unit, CRIUGM, Université de Montréal, Montréal, QC, Canada

17 HeartWise.ai Laboratory, Montreal Heart Institute, Montréal, QC, Canada

**Corresponding Author:**

Robert Avram, MD, MSc

Division of Cardiology, Department of Medicine, Director of HeartWise.ai

Montreal Heart Institute, University of Montreal

5000 Belanger Street, QC, Canada, H1T 1C8

robert.avram.md@gmail.com

**Competing Interests:** Robert Avram, MD, MSc: Spiralis Medical Inc. (stock option), Divoco AI (stock), FrontRx (co-founder and equity holder), Abbott (speaker fee), Boston Scientific (speaker fee), Boehringer Ingelheim (speaker fee), Novartis (speaker fee), Bristol Myers Squibb (investigator-initiated research grant), co-inventor on pending patent application WO 2022/261641 A1 (Method and System for Automated Analysis of Coronary Angiograms). Gordon J. Harris, PhD: Yunu (co-founder and equity holder), QuiqVu (co-founder and equity holder), Fovia (scientific advisory board). Rob Lewis: Radical Imaging (founder and owner), Adva Health Solutions (board member and shareholder). The other authors declare no competing interests.

Author Contributions: R.A. conceived the platform and the article, led the prospective deployments, and wrote the first draft. S.K. co-led the project, coordinated the radiology, neurology, and radiation-oncology use cases, and contributed Figure 1. P.T.-L. led the CathEF prospective evaluation and supervised the thoracic aortic aneurysm use case. J.K. contributed the thoracic aortic aneurysm use case and the platform deployment screenshots. J.C.-A. contributed the spinal cord imaging use cases. L.L.-G. contributed the neuroradiology use cases and reviewed the platform description. J.L.J. and A.M. contributed the pediatric musculoskeletal ultrasound use case. H.B. contributed the radiation-oncology use case. J.G.H. contributed the fairness and responsible-deployment content. P.-L.D. led the privacy impact assessment and wrote the privacy and data-governance section. G.J.H. and R.L. contributed the OHIF and Cornerstone platform foundations. All authors reviewed and approved the final manuscript.

**Ethics:** The prospective clinical evaluations of CathEF and DeepRV were conducted with research ethics board approval at the participating institutions, as reported in their primary publications. The platform usage and clinician-feedback metrics reported here are aggregate operational data from routine monitoring of PACS-AI installations and contain no patient-identifiable information.

**Data Availability:** The aggregate deployment and feedback data reported in this Comment (study and job counts, completion and failure causes, inference times, and structured clinician rating summaries) are available from the corresponding author upon reasonable request. The underlying imaging studies and patient-level data are not publicly available because they contain protected health information and remain within each hospital's environment under institutional policies and provincial privacy law. Data supporting the prospective CathEF and DeepRV evaluations are described in their respective primary publications.

**Code Availability:** The complete PACS-AI framework, including a skill that enables an agentic AI to migrate a model onto PACS-AI, is available at https://github.com/HeartWise-AI/pacs-ai-backend. A demonstrator is available at https://www.pacsai.co. OHIF source code is available at https://github.com/ohif and https://github.com/cornerstonejs.

**Acknowledgements:** We thank David Harris-Koblin (CIFAR) for coordinating the collaboration behind this article. This work was supported by a CIFAR network grant for AI for Health Imaging. Robert Avram is a Junior 2 clinical research scholar of the Fonds de Recherche du Québec-Santé (grant #380127). Houda Bahig is a Junior 2 clinical research scholar of the Fonds de Recherche du Québec-Santé (grant #349871). Jacob Jaremko is provided protected research time by Medical Imaging Consultants and is supported by CIFAR. Julie G. Hussin holds a Tier 2 Canada Research Chair from NSERC in Responsible Multi-omics Data Science. Pierre-Luc Déziel holds the Canada Research Chair on Health Data Protection and Valorization. Gordon Harris is Principal Investigator on US National Institutes of Health grants funding OHIF (5U24CA258511 and 1R03LM014994). Laurent Létourneau-Guillon is supported by a Fonds de Recherche du Québec-Santé (FRQ-S)/Fondation de l'Association des Radiologistes du Québec (FARQ) Junior 2 salary award (DOI: 10.69777/311203).

**Abstract**

We describe deploying imaging AI at six hospitals through PACS-AI, an open self-hosted platform. The binding constraint is not model accuracy but infrastructure to route studies, display results, capture feedback, and audit what runs. At one center, angiography models completed 515 of 607 jobs (84.8%); failures reflected absent diagnostic views, and 78.1% of 638 clinician ratings were positive. Publishing honest readiness levels for every model is itself a governance practice.

**Introduction**

Most clinical AI is reported as a single task-specific model, evaluated on one retrospective dataset and summarized in one performance table, with no durable integration into a clinical workflow and no mechanism for feedback, monitoring, or improvement after publication.[1] The gap between benchmark performance and clinical value is not primarily a modeling problem. Models trained on historical data degrade silently when the populations, scanners, acquisition protocols, and clinical behavior around them evolve, a phenomenon known as dataset shift.[2] Deployed tools that performed well in development may have missed many of the events they were meant to catch and generated large alert burdens in practice.[3] Recent guidance treats deployment as an ongoing, monitored process: silent-mode evaluation before a model is used by clinicians,[4] integration into the clinical pathway rather than a research folder, and a feedback loop that captures clinician agreement, correction, and outcomes.

Acute coronary syndrome (ACS) is a case in point: left-ventricular ejection fraction (LVEF) drives management, but echocardiography is often not available during coronary angiography and left ventriculography adds contrast and procedural burden, so a decision-quality estimate is needed in the cath lab, not hours or days later.[5]

We built PACS-AI, an open-source, self-hosted platform that lets a hospital deploy imaging AI inside routine care rather than alongside it. It ingests studies from the hospital PACS, runs containerized models on-site, returns results in the DICOM viewer clinicians are familiar with, and logs structured clinician feedback against a versioned model registry. It is currently installed at six hospitals (Montreal Heart Institute, Ottawa Heart Institute, University of Alberta, Polytechnique Montreal, University Health Network and Centre Hospitalier Universitaire de Montreal), and we report here the lessons learned running it. The platform works across any medical image: it carries model families across angiography, echocardiography, electrocardiography, radiography, and cross-sectional imaging (Table 1). We illustrate it with the cardiology deployment that motivated it, at the Montreal Heart Institute, where three models are deployed passively on all angiograms: CathEF, which estimates LVEF; DeepRV, which grades right-ventricular (RV) systolic function; and DeepCORO-CLIP, our angiogram foundation model for multi-view interpretation and stenosis estimation. CathEF[6] was evaluated prospectively in the catheterization laboratory[5] through PACS-AI, from silent-mode evaluation to clinical display to monitored use. The code is released open source, including a migration tool that lets an agentic AI port a new model onto the platform, with a public demonstrator at pacsai.co.[7]

**The PACS-AI platform**

PACS-AI[7] is a clinical operating system for running imaging AI applications: it runs several models side by side and provides standardized inputs and outputs. It is built as a fork of the Open Health Imaging Foundation (OHIF) zero-footprint DICOM web viewer, extended with an AI service tier, a model registry, and a structured-feedback interface. Clinicians therefore review AI outputs in a viewer they are already familiar with. Studies are ingested from the hospital picture archiving and communication system (PACS) and stored locally through an Orthanc DICOM server either upon physician initiation or automatically on a time-schedule,[8] so imaging data and model weights stay inside the hospital environment (Figure 1).

A scheduled control panel polls the PACS at operator-defined intervals, usually every 60 seconds, identifies eligible studies from their DICOM metadata, retrieves them into local Orthanc once acquisition is complete, and dispatches the containerized models that configurable rules select for that study by modality, site, care pathway, and consent; the browser receives those state changes over server-sent events instead of polling the platform. Inference can run during acquisition, after upload, in batch across an archive, or in silent mode before any clinician sees an output. Results are written back to the PACS as a structured report embedded in the DICOM, as a DICOM Segmentation object, or as a DICOM RTSTRUCT contour set for radiation-oncology planning, so the archive itself becomes the delivery channel.

In comparison, open imaging-AI frameworks such as MONAI Deploy[9], XNAT[10], and Kaapana[11] provide extensible infrastructure for DICOM integration, image viewing, containerized model execution, workflow orchestration, and, in some cases, model training and federated learning, but do not natively provide an integrated prospective clinical learning loop linking deployment, real-world outcomes, continuous model evaluation, and iterative improvement. Commercial orchestration platforms such as Aidoc aiOS, CARPL.ai, and Blackford provide multi-model deployment and monitoring but remain proprietary and are predominantly focused on radiology-specific workflows.[12] Relative to a vanilla Orthanc server with the OHIF viewer plug-in, PACS-AI adds the versioned model registry with model facts labels[13], rule-based routing and dispatch of containerized models based on settings such as frequency, metadata fields and dates, standardized manifests for inputs and outputs, silent-mode evaluation, structured clinician feedback, drift monitoring, and audit logging, as one open-source stack. A comparison of these platforms is provided in Supplementary Table S2.

Porting a new model does not require writing platform code. Each model is declared in a standardized manifest specifying modality, eligibility rules, container image, inputs, and output schema, alongside its Model Facts label. The input can be a 2D X-ray, a 3D CT scan or MRI, or video-based modalities such as transthoracic echocardiogram. The output is either a structured PDF, a JSON output or an application programming interface (API) command for integration with the electronic health record. Moreover, we release an AI agent skill, a structured instruction set that a large language model reads to generate the manifest and adapter, making migration of any model a supervised generation-and-review task.

**Two prospective deployment use-case studies evaluating impact in the field**

The active angiography models, DeepCORO-CLIP, CathEF, and DeepRV, processed 413 distinct prospective angiogram studies between June 6 and July 22, 2026 at the Montreal Heart Institute (MHI), and generated 607 study-by-model jobs, of which 515 completed and 92 failed (15.2%), at a throughput of 13 to 34 distinct studies per day and a mean inference time of 18.2 ± 6.1 s for DeepCORO-CLIP and 4.2 ± 2.1 s for CathEF and DeepRV. All 92 failures in this window shared a single upstream cause, a study retrieved into local Orthanc that contained no series matching the model's target modality, which is diagnostic coronary angiography, so dispatch was rejected before inference. Ingestion is automatic, so these counts cover every eligible study; clinical consultation of the results is the narrower measure. The Montreal Heart Institute installation has 25 physician and fellows' users, 80% of whom remain active, and the number of procedures in which clinicians consulted a model result rose from roughly 50 to roughly 110 per month over six months of use.

CathEF estimates LVEF directly from routine coronary angiography video.[6] It was evaluated prospectively through PACS-AI in an ACS cohort of 207 patients across two Canadian tertiary cardiac centres (MHI and Ottawa Heart Institute), with transthoracic echocardiography within seven days as the reference standard.[5] The model discriminated reduced systolic function with an AUROC of 0.84 for LVEF ≤50% and 0.90 for LVEF ≤40%, and showed reduced performance in right-coronary-artery culprit lesions and when the reference echocardiogram was delayed beyond two days.[5]

DeepRV, the second prospective deployment, extends the same substrate to right-ventricular systolic function, prognostically important yet not routinely quantified from angiography. Prospectively deployed in 82 STEMI cases, it discriminated reduced RV function with an AUROC of 0.83 and returned a result in a median of 5.1 seconds, fast enough to be used within the procedure; AI assistance improved reader accuracy from 72.1% to 77.6% among cardiologists and from 43.5% to 64.0% among medical students.[9] Because DeepRV runs on the video already acquired for CathEF, the two demonstrate the platform's multi-model core: a single acquisition drives two containerized inferences that return structured results in the same viewer. DeepCORO-CLIP adds multi-view interpretation and stenosis estimation on the same videos.[14]

Deployed this way, models run at the moment of decision, where their limitations become measurable. Outputs populate real-time worklists that flag likely reduced LVEF or RV function for early echocardiography and therapy review, and route low-confidence or discordant cases for adjudication. The worklist is live: each row carries a study's lifecycle and outcome state and the per-model progress of every dispatched job, streamed over server-sent events, and a model that did not run carries an explicit skip or failure reason, not a silent gap (Figure 3a). Those prospective deployments also happen at the Ottawa Heart Institute, whereas the four other centers are mainly using it for retrospective analysis now so no usage data is available.

**Monitoring, governance and privacy**

PACS-AI converts static model deployment into a continuously monitored learning loop. Every model in the registry ships with a versioned model facts label covering intended use, eligible population, performance overall and by subgroup, failure modes, calibration, and last-revalidation date, and it cannot reach clinical display without one.[15] Clinician feedback is structured and low-friction: agree or disagree, corrected label, uncertainty reason, image-quality flag, and downstream action. Readers rate any inference output at review, and a negative rating opens a fixed reason questionnaire.

Over a 13-week period ending August 3, 2026, at the MHI installation, 638 ratings covered 14.9% of returned reports, 498 of them (78.1%) positive, with weekly approval improving over the period. The 140 negative ratings cited a missed clinically relevant finding (41), measurement disagreement (33), unclear or verbose text (24), wrong series or laterality (19), latency (14), or other (9) (Figure 3b).

Before a model reaches clinical display, it undergoes silent-mode validation.[4] Every inference logs inputs, outputs, timestamps, model version, and clinician interaction, so input-data drift can be monitored continuously, not only outcome performance.[16] These logs are the prospective inputs that reporting guidelines for clinical AI evaluation now require: DECIDE-AI, for early clinical evaluation of decision-support systems,[17] TRIPOD+AI, for prediction models,[18] and other prospective-evaluation reporting standards. Versioned labels with pre-specified update criteria are the artifacts requested by the FDA's Predetermined Change Control Plan guidance[19]. The overall design is consistent with the FUTURE-AI international consensus on trustworthy medical AI and with American Heart Association and European Society of Cardiology calls for implementation science.[20]

Fairness is treated as a versioned, continuously recomputed registry property, not a one-time table. Each label reports performance and calibration by age, sex, race or ethnicity, and acquisition protocol, and the feedback loop tracks performance gaps, calibration drift, and representation coverage to detect emerging bias after deployment. A breach triggers silent-mode revalidation and, if unresolved, rollback, consistent with published strategies for mitigating bias in cardiovascular AI.[21] Moreover, during platform onboarding users undergo a fairness training module introducing them to the concept of fairness, its importance and how it can be measured and impact model predictions.

Because the platform processes identifiable imaging inside the hospital, monitoring is paired with a formal privacy framework. A privacy impact assessment under Québec's Act respecting health and social services information, which is more prescriptive on health data than GDPR or HIPAA, found no structural barrier to compliance: the platform runs on-premises and does not export data, it analyzes images patients already consented to acquire, and its access and retention settings can be tuned to local law at each site.[22] The assessment yielded seven governance practices we now apply at each installation, listed in Supplementary Note S1.

**Beyond angiography**

PACS-AI supports additional outputs such as segmentation masks. Voxel-wise predictions return to the archive as DICOM Segmentation objects or RTSTRUCT contour sets, loaded by the viewer as editable overlays in axial, coronal, sagittal, and volume-rendered views (Figure 2D). Contour edits are logged against the model version as structured feedback, a more informative signal than an agree-or-disagree flag, and can be reused to retrain the model; because the output is standard DICOM it persists in the archive and is readable by any conformant viewer.

The same container-and-feedback path now runs beyond angiography: every model family below was integrated by an independent team and runs on PACS-AI today, but at the integration or retrospective readiness reported in Table 1, and none has completed prospective validation on the platform (Figure 2). MedGemma 1.5, an open medical vision-language model, runs as a Docker/API adapter that drafts chest-radiograph reports (Figure 2C)[23]; the adapter demonstrates that any locally hosted vision-language model can be plugged in, not that this model is ready for clinical use. A pediatric hip-ultrasound adapter returns the alpha angle and femoral head coverage for clinician sign-off (Figure 2A), following published hip-dysplasia tools.[24] Neuroradiology adapters cover demyelinating cord lesions on spinal MRI (Figure 2B) and intracerebral hemorrhage with perihematomal edema on head CT (Figure 2D). A radiation-oncology model, OPTIMA-OPC, entered prospective observational deployment in July 2026 (Supplementary Table S1). We also tested an echocardiography secondary-reporting workflow that routes studies to open echo foundation models such as EchoPrime[25] or PanEcho[26], generates a secondary report stored separately from the signed clinical report and compares the two for clinically relevant discordance (Table 1).

**Key lessons from deployment**

First, a complete audit of the registry and its repositories was performed for each model (Table 1). The gate caught defects that a capabilities list would have hidden: the echocardiography adapters could not produce clean builds because of missing model assets such as lacking view-classification (required to route the right view to the right model), one integration's Model Facts file listed the wrong modality, and several model checkpoints were absent. None of these model families have reached clinical display; each must resolve its defects and pass silent-mode evaluation first.

Second, day-to-day operations were harder than the model work. Every site needs its own PACS and network integration as well as access to a graphic card processing unit cluster (GPU) with at least 48 gb of RAM. Moreover the structured feedback can be challenging to obtain routinely: clinicians review flagged and discordant cases more often than concordant ones, hence agreement rates are not unbiased estimates of accuracy. Model performance varies by site, protocol, and population, so our estimates should not be assumed to transfer, and adoption measures from a single centre may not generalize. Most model families in the registry remain at integration or retrospective-validation readiness (Table 1); only CathEF and DeepRV have completed prospective deployment. Self-hosting and open licensing discharge none of the privacy, cybersecurity, ethics, or regulatory obligations. We report deployment feasibility and discrimination, not clinical outcomes.

Third, the deployment environment matters as much as the model. Reusing an open-source viewer (OHIF) and an open-source DICOM server (Orthanc) allows a validated angiography model to reach the point of decision without a bespoke application.

DeepRV was added as a second deployment through a configuration change, showcasing the versatility of this platform.

Fourth, a model should pass through standardized evaluation on the PACS-AI platform before it is released for clinical use, because prospective behavior can diverge sharply from retrospective metrics.[4] An installed base makes the intermediate step practical: the same container can run retrospectively across several hospitals' archives, under local governance and without data leaving any institution, surfacing site, protocol, and population effects before any clinician sees an output. Using versioned containers standardizes the harness around the actual model, which is crucial because we observed, for instance, discrepancies in performance between different PyTorch libraries, since certain PyTorch libraries are processing the images differently from others, leading to inconsistent results. to address this, we are shipping the model containers with 10 benchmark exams to prove that the model is able to replicate the prediction we had internally in each center before deployment. Silent mode then extends that assessment into live workflow.

Fifth, feedback must be structured and auditable to be useful. Free-text impressions cannot drive recalibration, fairness monitoring, or regulatory documentation; typed fields logged against model versions, on the other hand, translate quantitative measures into actionable steps (Figure 3b). That also means clinicians need training in a model's intended use, usually done during platform onboarding, and limits before their feedback means anything.

Across these deployments, the limiting factor was rarely model performance; it was the missing infrastructure to deploy, monitor, and correct models after publication. Prioritized worklists and single-task triage have already been proven at scale, but as closed products (Supplementary Table S2); what an open, self-hosted environment adds is the ability to break silos, to close the feedback loop where previous products are limited, and to state, model by model, what is actually running and what is not.

**PACS-AI demonstrates that successful imaging AI deployment depends not only on model performance, but also on reliable clinical integration, prospective validation, structured feedback, and continuous monitoring. Its open, self-hosted architecture provides a practical foundation for deploying and evaluating imaging AI across institutions, although its effect on clinical outcomes remains to be established.**

**Figures and Tables**

**Table 1.** PACS-AI model families and readiness level.

| **Model family** | **Modality** | **Primary task** | **Validation / readiness** | **Publication reference** |
|---|---|---|---|---|
| CathEF, CathAI, DeepCORO, DeepRV | XA coronary angiography | LVEF and RV function; anatomy, stenosis, coronary findings, SYNTAX scoring | Mixed R3–R4; internal/external validation; prospective deployment for CathEF and DeepRV | CathEF[6]; DeepRV[27] |
| EchoPrime, PanEcho | TTE ultrasound | Multi-view interpretation, measurements, diagnoses, report support | R3; published retrospective multisite validation; local prospective PACS-AI validation not established | EchoPrime[25] PanEcho[28] |
| DeepECG-SL, DeepECG-SSL | 12-lead ECG | Multilabel ECG interpretation and task-specific biomarkers | R3; external validation reported; prospective PACS-AI deployment not claimed | DeepECG[29] |
| MedGemma 1.5 | Chest radiograph | Prompted image-to-text interpretation | R1 integration demonstration; | Google MedGemm |
|  | y (DX/CR) | and draft report generation | benchmark evaluation is not local clinical validation | a 1.5 model card[23] |
| CIED-AI, BrainGPT, Hemorrhage nnU-Net, TotalSegmentator | CXR and CT | Device detection/classification; brain-CT reporting; hemorrhage and whole-body segmentation | Mostly R1–R3; public evidence for CIED-AI, BrainGPT, TotalSegmentator; no PACS-AI prospective validation found | CIED-AI (Lauzier et al., Heart Rhythm 2023); BrainGPT (Li et al., Nat Commun 2025) |
| Pediatric hip-ultrasound AI, MEDO-Hip workflow | Handheld hip ultrasound | DDH landmark segmentation, measurements, follow-up recommendation | External implementation evidence, not yet a shipped PACS-AI model; expected time saving unmeasured as yet | Alberta primary-care study[24] |
| TAA-AI, Automated Surveillance of TAA | Cardiac CT, CTA | Aortic trunk segmentation, centerline-based maximum diameter measurements on predefined | R1; internal evaluation in progress | Not published yet |

| | | landmarks. | | |
|---|---|---|---|---|
| AMO-ENE | CT | Head & neck treatment outcome prediction in low-risk patients | External implementation, integrated into PACS-AI; ongoing prospective trial | Hénique et al., arXiv 2026 |
| Spine segmentation | MRI, CT | Semantic segmentation of individual vertebrae and disc for subsequent morphometric analysis | R3; multi-site evaluation of spine morphometry measurements in lumbar pathologies (spinal canal stenosis, lateral recess, and foraminal stenosis) | SpineReport (Molinier et al., arXiv 2026) Spine segmentation (Warszawer et al., ISMRM 2024) |
| Multiple sclerosis spinal cord lesion segmentation | MRI | Segmentation of spinal cord MS lesions | R3; multi-site evaluation of model performance | Generalizable Multiple Sclerosis Segmentation model [30] |

Legend: Readiness levels: R1, integration demonstration; R2, internally validated model; R3, externally validated model; R4, prospectively deployed.

**Abbreviations:** CIED, cardiac implantable electronic device; CT, computed tomography; CTA, computed tomography angiography; CXR, chest radiograph; DDH, developmental dysplasia of the hip; DX/CR, digital radiography / computed radiography; ECG, electrocardiogram; LVEF, left-ventricular ejection fraction; MRI, magnetic resonance imaging; nnU-Net, no-new-U-Net; PACS, picture archiving and communication system; RV, right ventricular; SYNTAX, Synergy between Percutaneous Coronary Intervention with Taxus and Cardiac Surgery; TAA, thoracic aortic aneurysm; TTE, transthoracic echocardiography; XA, X-ray angiography.

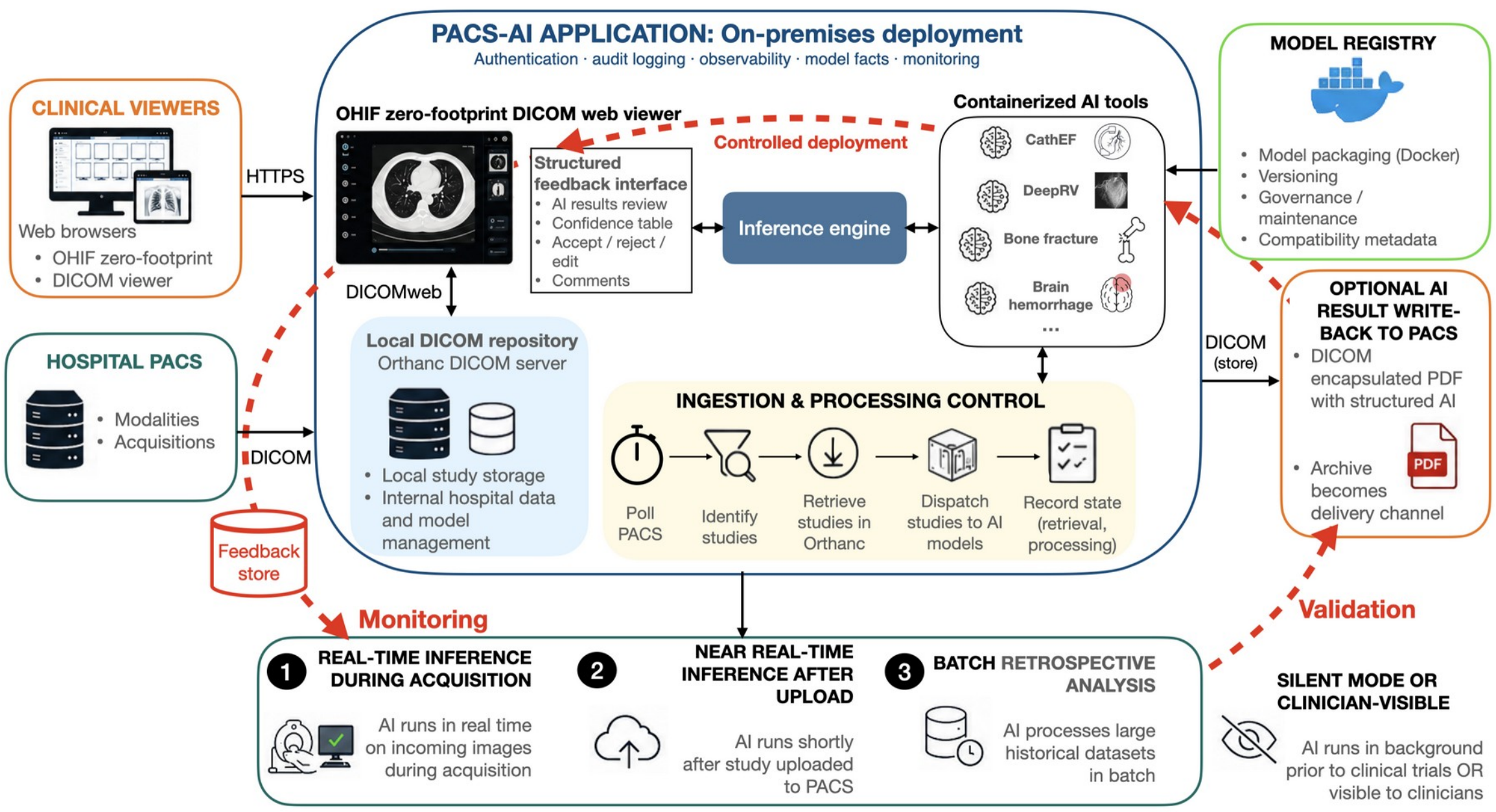

**Figure 1.** PACS-AI architecture and data flow, from the hospital PACS through scheduled ingestion and the local Orthanc store to the containerized models in the registry, with results returned to the OHIF-fork viewer as AI overlays and structured clinician feedback logged to the audit database.

**Figure 2. Four inference applications deployed in PACS-AI**

**A** **Pediatric hip ultrasound — automated hip dysplasia measurement**

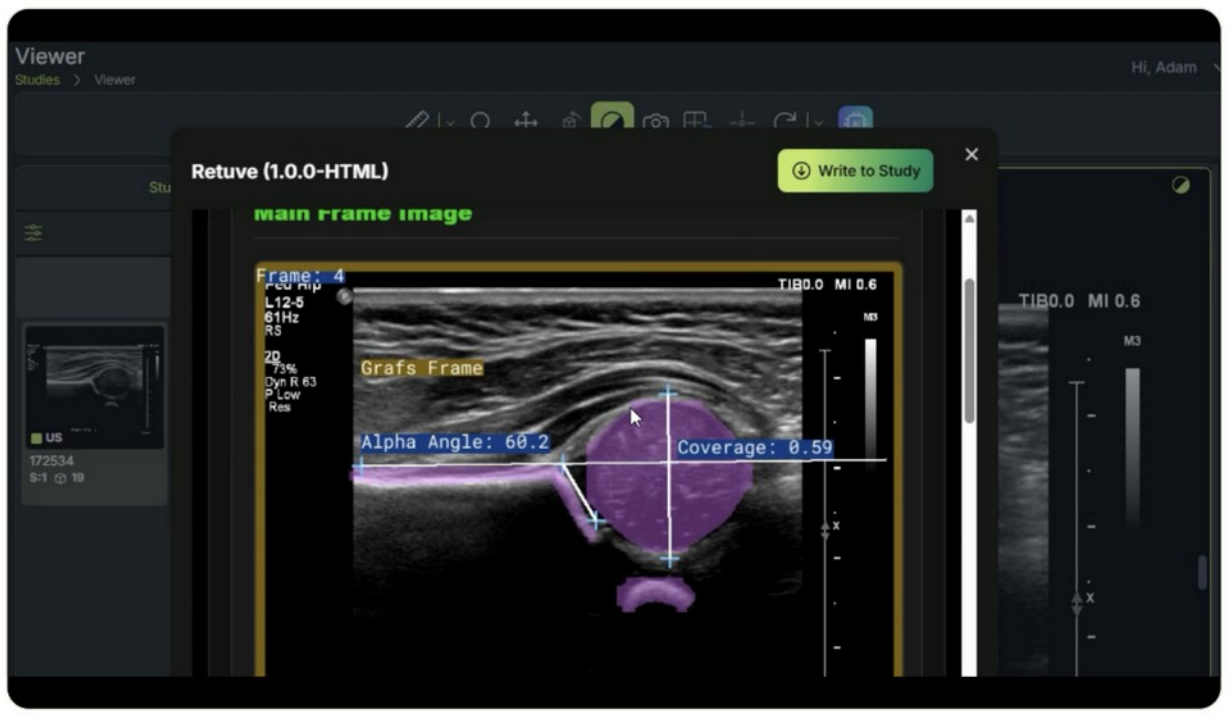

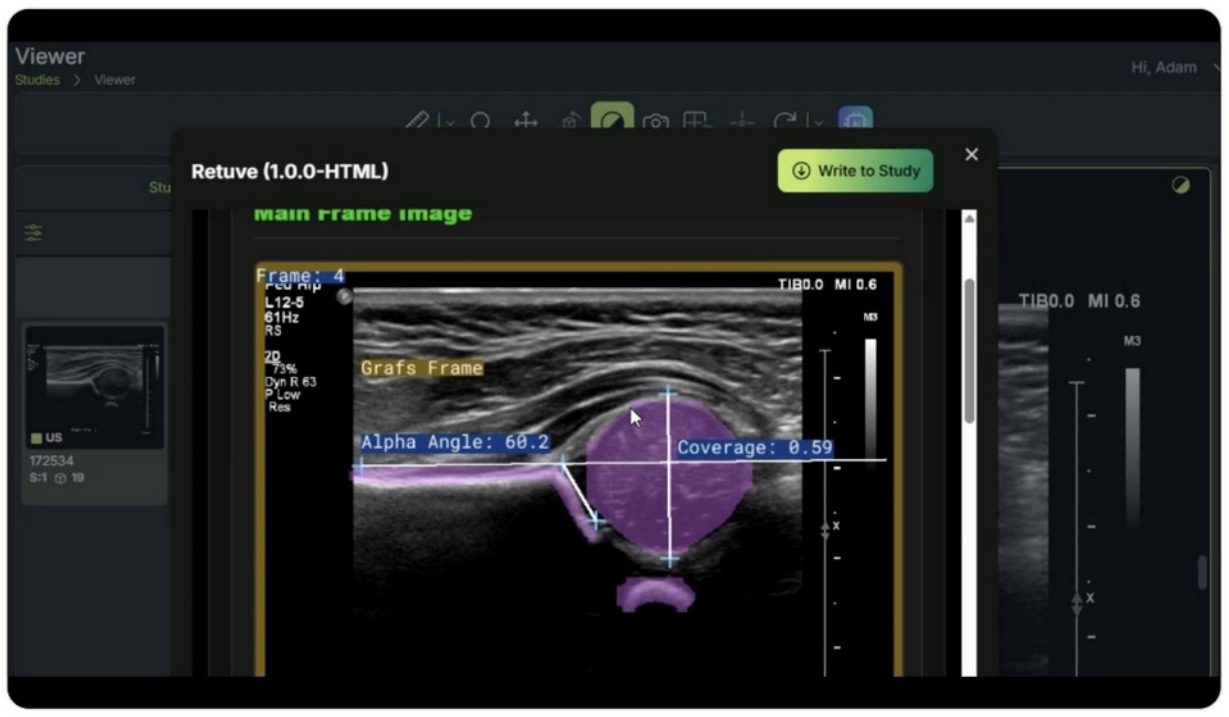

**B** **Demyelinating lesion detection on spinal cord MRI**

**C** **Vision–language reporting on chest radiography using MedGemma**

**D** **Intracerebral hemorrhage and perihematomal edema segmentation**

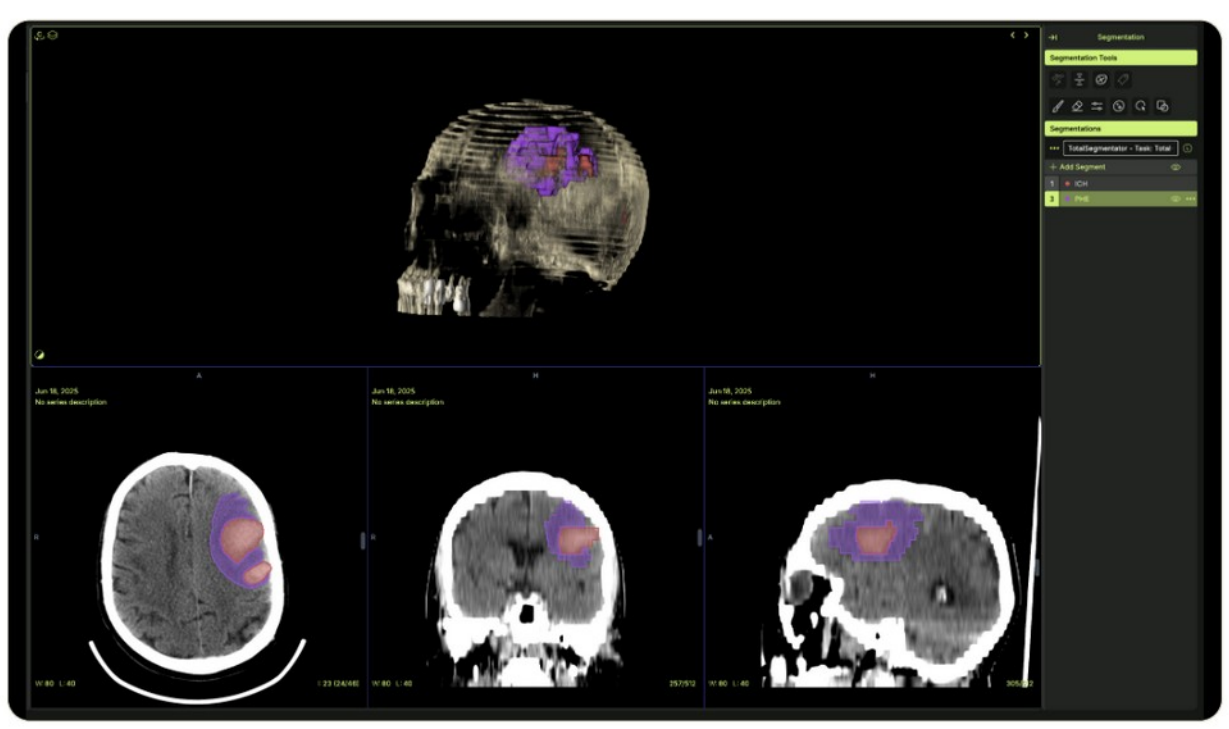

**Legend**. **(A)** Pediatric hip ultrasound: Retuve (v1.0.0) selects the Graf reference frame from a cine sweep, segments the ilium, femoral head and labrum, and returns an alpha angle of 60.2° with a femoral head coverage of 0.59. **(B)** Sagittal T1 MPR of the cervical spine with candidate demyelinating cord lesions rendered as a segmentation overlay on the native series. **(C)** MedGemma 1.5 returns a structured narrative description of a chest radiograph (heart size, lung fields, pleural spaces, and diaphragm), which the reader can write back into the study; the interface is shown in French. **(D)** Non-contrast head CT with two labels, hematoma (red) and perihematomal edema (purple), displayed on the volume rendering and on the axial, coronal and sagittal planes (W 80, L 40). Each application runs as a self-contained inference plugin invoked from the viewer toolbar and returns results as DICOM segmentations or structured HTML. Panels are unmodified screen captures.

**Figure 3. Operational surface of PACS-AI: real-time study worklist and model-feedback dashboard, shown for the Centre Hospitalier de l'Université de Montréal (CHUM) installation.**

(A)

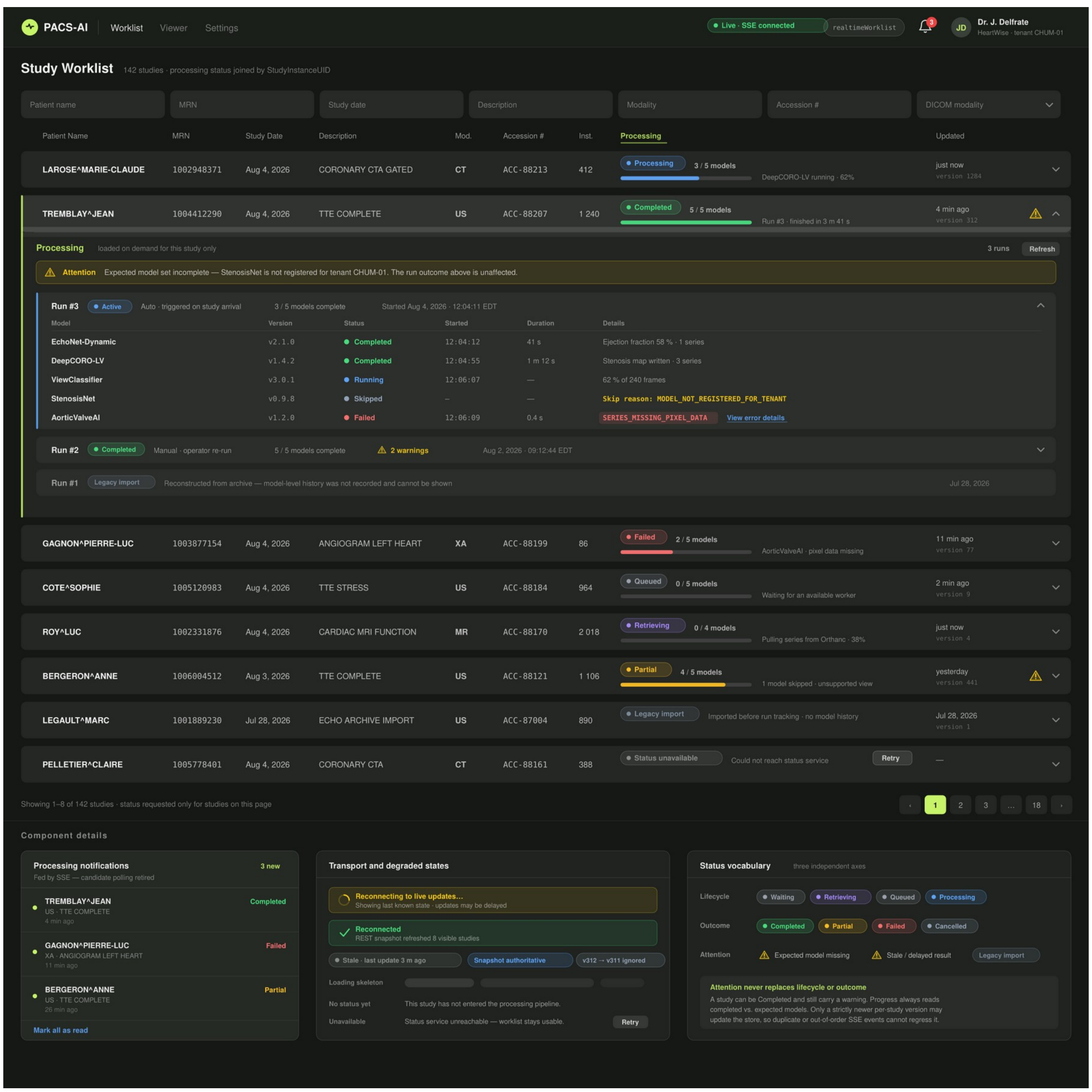
PACS-AI
Worklist
Viewer
Settings
Live · SSE connected
realtimeWorklist
3
JD
Dr. J. Delfrate
HeartWise · tenant CHUM-01
Study Worklist
142 studies · processing status joined by StudyInstanceUID
Patient name
MRN
Study date
Description
Modality
Accession #
DICOM modality
Patient Name
MRN
Study Date
Description
Mod.
Accession #
Inst.
Processing
Updated
LAROSE^MARIE-CLAUDE 1002948371 Aug 4, 2026 CORONARY CTA GATED CT ACC-88213 412
Processing 3 / 5 models
DeepCORO-LV running · 62%
just now
version 1284
TREMBLAY^JEAN 1004412290 Aug 4, 2026 TTE COMPLETE US ACC-88207 1 240
Completed 5 / 5 models
Run #3 · finished in 3 m 41 s
4 min ago
version 312
Processing
loaded on demand for this study only
3 runs
Refresh
Attention Expected model set incomplete — StenosisNet is not registered for tenant CHUM-01. The run outcome above is unaffected.
Run #3 Active Auto · triggered on study arrival 3 / 5 models complete Started Aug 4, 2026 · 12:04:11 EDT
Model Version Status Started Duration Details
EchoNet-Dynamic v2.1.0 Completed 12:04:12 41 s Ejection fraction 58 % · 1 series
DeepCORO-LV v1.4.2 Completed 12:04:55 1 m 12 s Stenosis map written · 3 series
ViewClassifier v3.0.1 Running 12:06:07 — 62 % of 240 frames
StenosisNet v0.9.8 Skipped – — Skip reason: MODEL_NOT_REGISTERED_FOR_TENANT
AorticValveAI v1.2.0 Failed 12:06:09 0.4 s SERIES_MISSING_PIXEL_DATA View error details
Run #2 Completed Manual · operator re-run 5 / 5 models complete 2 warnings Aug 2, 2026 · 09:12:44 EDT
Run #1 Legacy import Reconstructed from archive — model-level history was not recorded and cannot be shown Jul 28, 2026
GAGNON^PIERRE-LUC 1003877154 Aug 4, 2026 ANGIOGRAM LEFT HEART XA ACC-88199 86
Failed 2 / 5 models
AorticValveAI · pixel data missing
11 min ago
version 77
COTE^SOPHIE 1005120983 Aug 4, 2026 TTE STRESS US ACC-88184 964
Queued 0 / 5 models
Waiting for an available worker
2 min ago
version 9
ROY^LUC 1002331876 Aug 4, 2026 CARDIAC MRI FUNCTION MR ACC-88170 2 018
Retrieving 0 / 4 models
Pulling series from Orthanc · 38%
just now
version 4
BERGERON^ANNE 1006004512 Aug 3, 2026 TTE COMPLETE US ACC-88121 1 106
Partial 4 / 5 models
1 model skipped · unsupported view
yesterday
version 441
LEGAULT^MARC 1001889230 Jul 28, 2026 ECHO ARCHIVE IMPORT US ACC-87004 890
Legacy import Imported before run tracking · no model history
Jul 28, 2026
version 1
PELLETIER^CLAIRE 1005778401 Aug 4, 2026 CORONARY CTA CT ACC-88161 388
Status unavailable Could not reach status service
Retry
—
Showing 1–8 of 142 studies · status requested only for studies on this page
1 2 3 … 18
Component details
Processing notifications
3 new
Fed by SSE — candidate polling retired
TREMBLAY^JEAN
US · TTE COMPLETE
4 min ago
Completed
GAGNON^PIERRE-LUC
XA · ANGIOGRAM LEFT HEART
11 min ago
Failed
BERGERON^ANNE
US · TTE COMPLETE
26 min ago
Partial
Mark all as read
Transport and degraded states
Reconnecting to live updates...
Showing last known state · updates may be delayed
Reconnected
REST snapshot refreshed 8 visible studies
Stale · last update 3 m ago
Snapshot authoritative
v312 → v311 ignored
Loading skeleton
No status yet
This study has not entered the processing pipeline.
Unavailable
Status service unreachable — worklist stays usable.
Retry
Status vocabulary
three independent axes
Lifecycle
Waiting
Retrieving
Queued
Processing
Outcome
Completed
Partial
Failed
Cancelled
Attention
Expected model missing
Stale / delayed result
Legacy import
Attention never replaces lifecycle or outcome
A study can be Completed and still carry a warning. Progress always reads
completed vs. expected models. Only a strictly newer per-study version may
update the store, so duplicate or out-of-order SSE events cannot regress it.

(B)

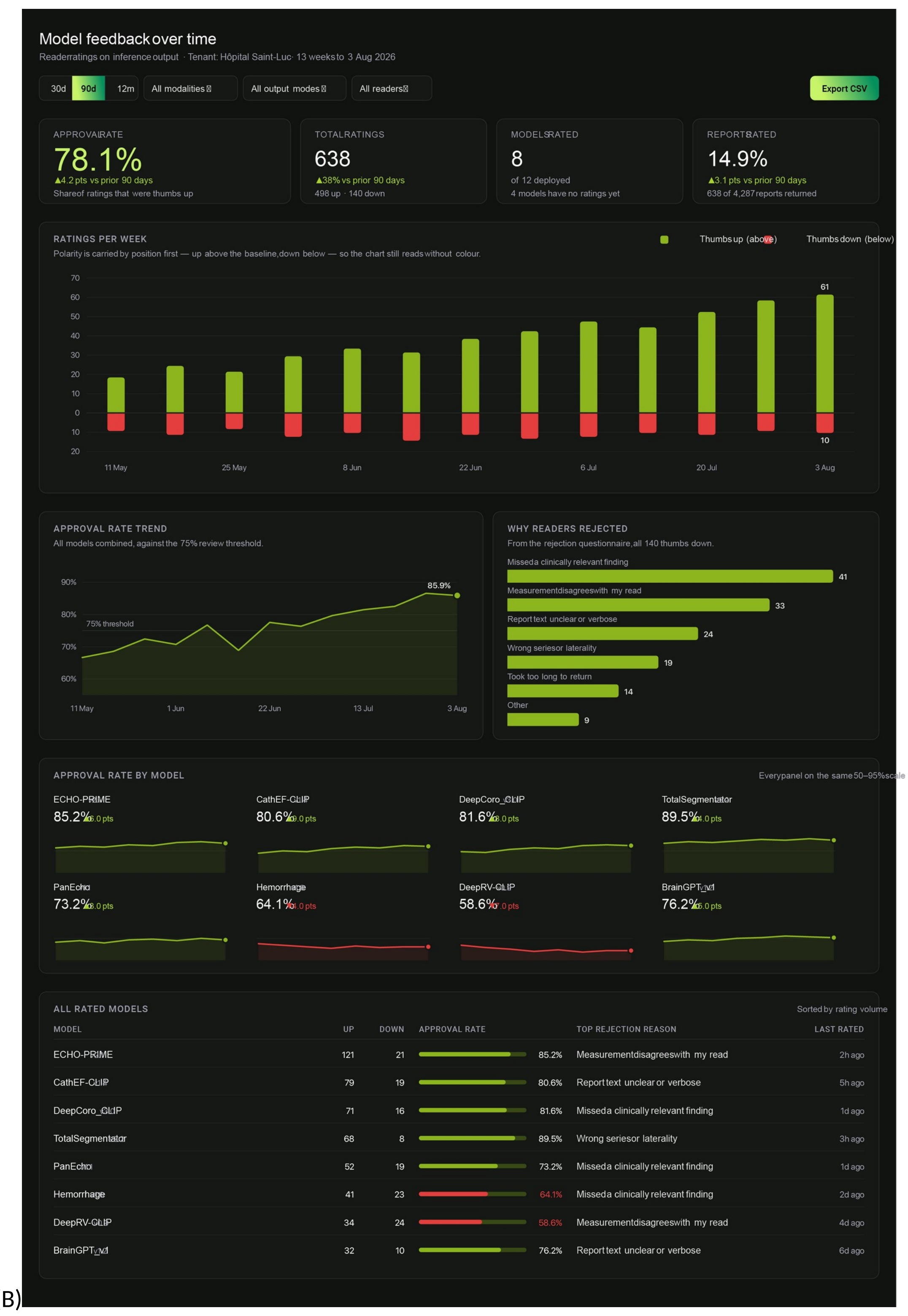

(B)

**Legend. (a)** Real-time worklist. Each row is one study retrieved from the hospital PACS, with its lifecycle state and the progress of every dispatched model job; state changes stream to the browser over server-sent events, so results appear as inference proceeds.

Expanding a study opens its run history, which records for every model whether the run was triggered automatically or manually, the pinned model version, start time, duration, and outcome. A model that did not run carries an explicit skip or failure reason, separating a skip (not registered for that site, unsupported view) from a container failure (missing pixel data, out of memory), with a link to the underlying logs. **(b)** Model-feedback dashboard for the 13 weeks to August 3, 2026, at the MHI installation, across all modalities and registered models. Readers rate any inference output at the point of review; a negative rating opens a fixed questionnaire recording the reason. Weekly rating volume is shown above and below the baseline, and approval rate is tracked against a configurable review threshold that triggers re-examination. Ratings from models not in clinical use were collected in silent mode. Patient identifiers in panel a are de-identified.

## Supplementary Materials

### Supplementary Note S1. Governance practices

The privacy impact assessment under Québec's Act respecting health and social services information yielded seven governance practices, applied at each installation:

1 Declare AI use in the imaging consent form.

2 Host and authenticate the platform on hospital infrastructure.

3 Log who accessed what, when, and why.

4 Set explicit retention and destruction schedules.

5 Document model-selection criteria and treat alerts as prioritizations, never verdicts.

6 Verify the configuration against local law before go-live.

7 Where feasible, commission a site-specific privacy impact assessment.

**Supplementary Table S1. Prospective performance of CathEF, DeepRV, and OPTIMA-OPC.**

| Model | Task and threshold | Development / validation cohort | Reference standard | AUROC | Key limitations |
|---|---|---|---|---|---|
| CathEF | LVEF ≤50% and ≤40% from coronary angiography | Prospective ACS cohort, 207 patients, 2 Canadian tertiary centres[5] | Transthoracic echocardiography within 7 days | 0.84 (≤50%); 0.90 (≤40%) | Reduced in RCA culprit lesions and with delayed reference echocardiography |
| DeepRV | Right-ventricular systolic function from coronary angiograms | 8,053 studies / 6,923 patients (MHI); external 2,247 studies (UCSF); prospective 82 STEMI-PCI cases[27] | Expert echocardiography-based RV grade | 0.80 internal; 0.75 external; 0.83 prospective | High NPV (95.8%) but low PPV (27.7%), supporting rule-out rather than rule-in use; AUROC 0.55 in RCA-culprit STEMI and 0.65 in STEMI overall; echocardiographic reference standard only, without core-lab adjudication |
| OPTIMA-OPC | Recurrence-risk stratificati | 1,428 patients, two | Locoregional or distant recurrence on | 0.796 internal; 0.770 | Retrospective development; prospective |
|  | on in HPV+ OPC | Canadian centres. Development 1,075; external validation 353; prospective observational deployment from July 2026 | institutional follow-up | external; 0.830 with PET added; 0.741 for clinical variables alone | performance using PACS-AI ongoing |

**Abbreviations:** ACS, acute coronary syndrome; AUROC, area under the receiver operating characteristic curve; HPV, human papillomavirus; LVEF, left-ventricular ejection fraction; MHI, Montreal Heart Institute; OPC, oropharyngeal carcinoma; PCI, percutaneous coronary intervention; PET, positron emission tomography; RCA, right coronary artery; RV, right ventricular; STEMI, ST-elevation myocardial infarction; UCSF, University of California, San Francisco.

**Supplementary Figure S1. Case timeline for CathEF and DeepRV deployed inside the clinical PACS viewer.**

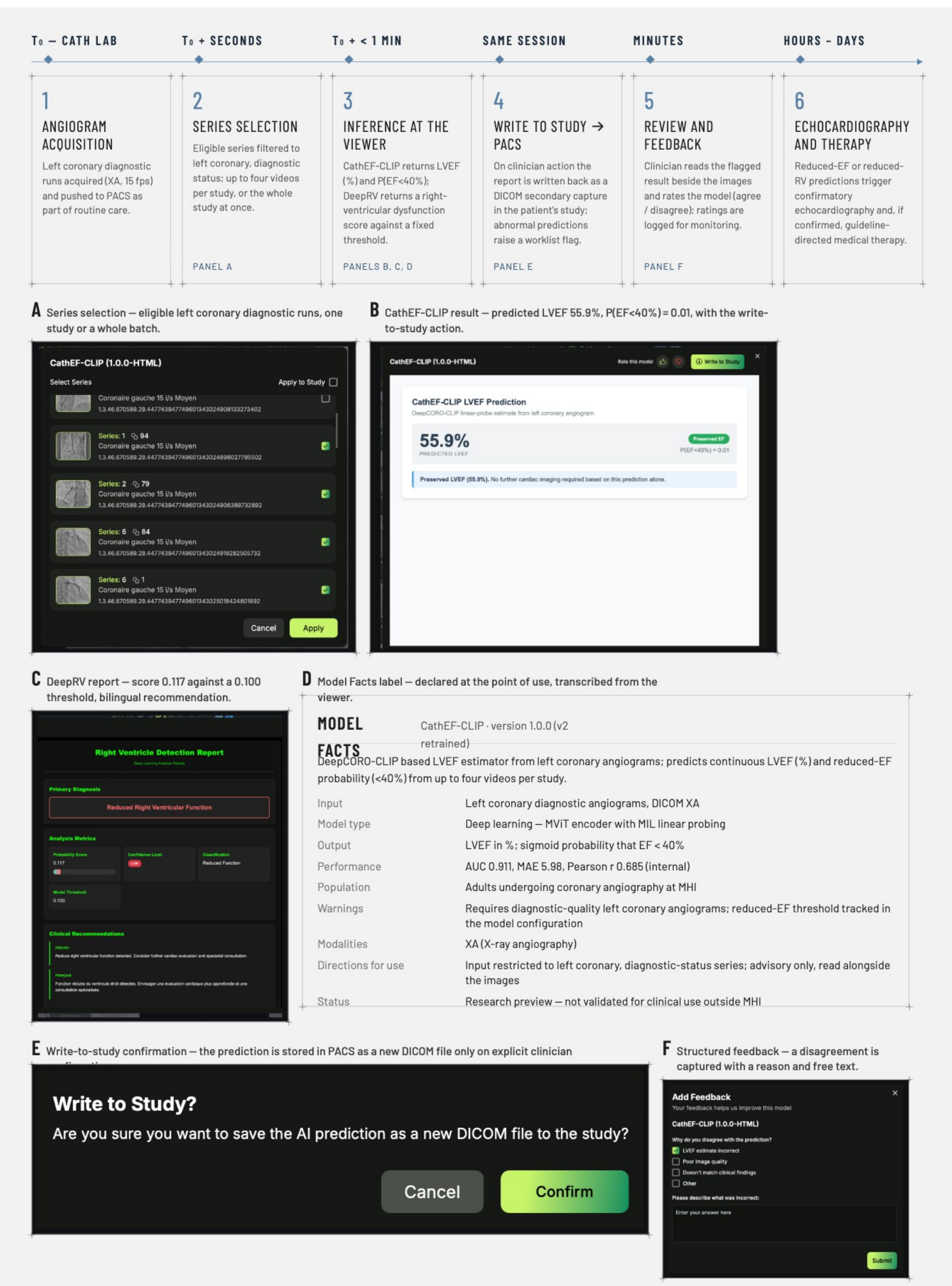

**Legend:** Diagnostic left coronary angiograms acquired in the cath lab (1) are filtered to eligible series in the viewer (2, panel A); inference runs on the selected videos and returns a predicted LVEF with a reduced-EF probability, or a right-ventricular dysfunction score (3, panels B and C), each accompanied by a Model Facts label declaring inputs, architecture, internal performance and use restrictions (panel D). On clinician action the result is written back to the patient's study in PACS as a DICOM secondary capture, and abnormal predictions raise a worklist flag (4). The clinician reviews the flagged result alongside the images and rates the model, logging feedback for monitoring (5; Figure 3b). Abnormal predictions route the patient to confirmatory echocardiography and, where confirmed, guideline-directed therapy (6). All predictions are advisory; no automated action is taken without clinician review.

**Supplementary Table S2. Comparison of clinical imaging and cardiovascular signal-AI deployment platforms**

| Category | Representative products/projects | Access and deployment model | Scope | Post-deployment monitoring, feedback, and model updating |
|---|---|---|---|---|
| **Open imaging-AI infrastructure, data, and visualization frameworks** | MONAI Deploy, XNAT, Kaapana, OHIF | Open source; generally self-hostable | Heterogeneous building blocks rather than equivalent deployment platforms: MONAI Deploy packages and runs DICOM-capable AI applications; XNAT manages imaging data and containerized processing; Kaapana orchestrates workflows, annotation, training, and federated learning; OHIF is an extensible DICOMweb viewer and annotation framework. | Human-in-the-loop annotation and iterative training exist, notably in Kaapana and the MONAI ecosystem, and XNAT and OHIF can store clinician annotations. No turnkey cardiovascular loop linking point-of-care predictions, adjudication, outcomes, surveillance, and governed model updating is publicly documented. |

| Category | Representative products/projects | Access and deployment model | Scope | Post-deployment monitoring, feedback, and model updating |
|---|---|---|---|---|
| **Proprietary enterprise imaging-AI orchestrators and marketplaces** | Aidoc aiOS, CARPL.ai, Blackford Platform, deepcOS | Proprietary enterprise platforms | Multi-model, often multi-vendor orchestration spanning PACS/RIS/EHR integration, predeployment validation, analytics, governance, and lifecycle management. Predominantly imaging-centric, although Aidoc extends into cardiology. | **Drift detection and override tracking (Aidoc), ground-truth capture and real-time validation (CARPL), accept/reject review and 24/7 monitoring (Blackford), deployment monitoring and governance (deepc). An open, institution-controlled, outcome-linked retraining loop is not publicly documented.** |
| **Proprietary cardiovascular imaging and physiologic-analysis suites** | HeartFlow, Cleerly, Ultromics | Proprietary; regulated modules | Multi-analysis but disease- or modality-constrained: HeartFlow (FFRCT, RoadMap, plaque analysis, PCI planning), Cleerly (plaque, stenosis, and ischemia from CCTA), and Ultromics (echo-based HFpEF and amyloidosis). | Clinical registries and evidence-generation programs exist (NAVIGATE-PCI, INVICTUS, CONFIRM2), but these are evidence programs, not open, customer-controlled model-retraining infrastructure. |
| **Proprietary cardiovascular care-** | Viz.ai, Anumana, Eko | Proprietary; regulated algorithms | Multi-algorithm rather than single-task: Viz.ai spans >50 | Workflow analytics, care coordination, and downstream |

| Category | Representative products/projects | Access and deployment model | Scope | Post-deployment monitoring, feedback, and model updating |
|---|---|---|---|---|
| **coordination and signal-AI platforms** | | and enterprise software | cleared algorithms across CT, ECG, echo, and care coordination; Anumana has cleared ECG-AI algorithms for low ejection fraction, pulmonary hypertension, and amyloidosis; Eko combines ECG and heart sounds for low ejection fraction, atrial fibrillation, and murmurs. | pathway tracking exist in parts of this group, but an open, institution-controlled, cross-product model-updating loop is not publicly documented. |
| **PACS-AI** | CathEF, DeepRV, CathAI/DeepCORO, echocardiography models, and ECG models within the broader PACS-AI/DeepECG program | Open-source research core ; self-hosted/on-premises deployment | Vendor-agnostic, containerized local inference with PACS integration, integrated viewing, point-of-care deployment, and prospective validation across cardiovascular imaging and signals. | Captures clinician feedback natively and tracks predictions longitudinally, supporting continuous evaluation and governed periodic model updates. |